# Multi-Airport Delay Prediction with Transformers

Liya Wang[1], Alex Tien[2], and Jason Chou[3]
*The MITRE Corporation, McLean, VA, 22102, United States*

**Airport performance prediction with a reasonable look-ahead time is a challenging task and has been attempted by various prior research. Traffic, demand, weather, and traffic management actions are all critical inputs to any prediction model. In this paper, a novel approach based on Temporal Fusion Transformer (TFT) was proposed to predict departure and arrival delays simultaneously for multiple airports at once. This approach can capture complex temporal dynamics of the inputs known at the time of prediction and then forecast selected delay metrics up to four hours into the future. When dealing with weather inputs, a self-supervised learning (SSL) model was developed to encode high-dimensional weather data into a much lower-dimensional representation to make the training of TFT more efficiently and effectively. The initial results show that the TFT-based delay prediction model achieves satisfactory performance measured by smaller prediction errors on a testing dataset. In addition, the interpretability analysis of the model outputs identifies the important input factors for delay prediction. The proposed approach is expected to help air traffic managers or decision makers gain insights about traffic management actions on delay mitigation and once operationalized, provide enough lead time to plan for predicted performance degradation.**

## I. Nomenclature

| | | |
|---|---|---|
| $t$ | = | current time step |
| $k$ | = | number of time lags |
| $\tau$ | = | step from current time $t$, $\tau = [1, \ldots, \tau_{max}]$ |
| $\tau_{max}$ | = | maximum look ahead times |
| $X_t$ | = | exogenous time series at time $t$ |
| $F_t$ | = | known factors at time $t$ |
| $S_i$ | = | static variable $i$, which is invariant to time $t$ |
| $y_t$ | = | response variable at time $t$ |
| $mae$ | = | mean absolute error |
| $mse$ | = | mean squared error |

## II. Introduction

Flight delays have been a concern for multiple stakeholders such as passengers, airlines, airport authorities, and air navigation service providers. According to the Federal Aviation Administration (FAA) delays cost estimation [1], the current hourly cost to an airline ranges from about $1,400 to $4,500, with the value of passenger time ranging from $35 to $63 per hour. Flight delays costed US economy about 33 billion of dollars in 2019 [2]. To minimize the impact of flight delays, there is an operational need for developing a real-time delay monitoring and delay prediction system that can help airlines or aviation authorities proactively make delay mitigation strategies, thus reducing consequent economic losses and passenger frustration. Despite decades of research on flight delay prediction (e.g., [8], [9], [10], [11]), the challenges remains in reasonably modelling the correlation among delays, flight demand, system constraints due to weather, and traffic management actions. In particular, to incorporate weather data into modeling, traditional methods have to use rule-based aggregation methods, which may lose important spatiotemporal information for high-dimensional gridded weather data (e.g., Corridor Integrated Weather System (CIWS) [4], Rapid Update Cycle (RUC) wind [5], Himawari satellite image data [6]). These challenges present a research opportunity to use modern Artificial Intelligence (AI) approaches to build a model that accounts for complex temporal dynamics of the inputs known at the time of prediction and then forecast system performance into the future.

---

[1] Lead Artificial Intelligence Engineer, Department of Operational Performance
[2] Principal Engineer, Department of Operational Performance
[3] Lead Data Scientist, Department of Operational Performance

This paper proposed an AI-enabled approach to develop a multi-airport delay prediction model leveraging modern AI techniques: Self-Supervised learning (SSL) and Transformers [7]. SSL is used to learn useful representation from high-dimensional data, such as weather, and Transformer-based methods are good at using attention mechanisms to capture complex time dependencies between inputs and outputs (e.g., [12], [13], [14], [15]). To the best of our knowledge, no research has attempted to combine SSL technique and Transformers together to build a performance prediction system with consideration of gridded weather data, traffic data, and traffic management initiative (TMI) data together. Our research seeks to shed light on SSL and Transformers to solve aviation problems.

The remainder of this paper is organized as follows: Section III describes our problem formulations; and Section IV gives a short introduction about our input data sources used in the research. Section V provides details of the deep learning (DL) modeling process, and the results are shown in Section VI. Finally, we conclude and describe next steps in Section VII.

### III. Problem Formulation

Modeling the performance of multiple airports in a region (all at once) is necessary, as compared to modeling one airport at a time, since the current TFM decision-making process is driven by regional traffic characteristics and stakeholder's concerns. Our goal is to predict departure (i.e., wheels-off) delays and arrival delays for multiple airports over a strategic time horizon (4-8 hours). In a generic form, it can be formulated as a multi-variate, multi-step, time-series forecasting problem, as mathematically expressed in Eq. 1. Time lag variable $k$ is a parameter that represents the look-back time of what has happened, while the maximum look-ahead time $\tau_{max}$ is the parameter to be determined by the data updating frequency and prediction needs. Fig. 1 further illustrates the relationship of inputs and outputs in a given time horizon. The inputs are made up of several parts:
- Past prediction targets, which are observed delays up to prediction time;
- Observed inputs, such as actual arrival counts and departure counts;
- Known factors available for both past and future times, such as arrival and departure demand;
- Static variables, such as airport ID.

$$y_{t+\tau} = f(y_{t-1}, y_{t-2}, \ldots y_{t-k}, X_{t-1}, \ldots, X_{t-k}, F_{t-k}, \ldots, F_t, F_{t+1}, \ldots, F_{t+\tau}, S_1, S_{2,\ldots,}S_n) \quad (1)$$

where:
$t$ = current time step,
$k$ = number of time lags,
$\tau$ = step from current time $t$, $\tau = [1, \ldots, \tau_{max}]$,
$\tau_{max}$ = maximum look ahead times,
$X_t$ = observed inputs at time $t$,
$F_t$ = known factors at time $t$,
$S_i$ = static variable i, which is invariant to time $t$, and
$y_t$ = target variable at time $t$.

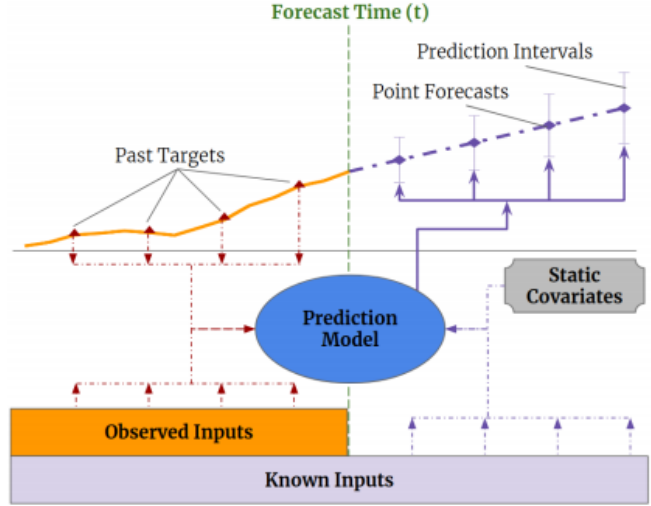

**Fig. 1 Illustration of multi-horizon forecasting [13].**

## IV.    Data Sources

We selected North-East corridor (NEC) as the study subject and developed a delay prediction model for four major airports in the region: John F. Kennedy International Airport (JFK), Newark Liberty International Airport (EWR), LaGuardia Airport (LGA), and Philadelphia International Airport (PHL). We identified three data sources that are relevant for the multi-airport delay prediction, concerning their acceptance in performance review and real-time availability. Table 1 lists the three datasets and summarizes the variables used for modeling.

**Table 1: Data Sources for Multi-Airport Performance Prediction**

| Category | Data Source | Variables and Definitions |
|---|---|---|
| **Weather** | Corridor Integrated Weather System (CIWS) | This gridded numerical weather data provides: Vertically Integrated Liquid (VIL) and Echo Top (ET) |
| **Traffic and Airport Performance** | Aviation System Performance Metrics (ASPM) | Traffic and performance data of airports. The data table with information such as:<br>• Airport ID<br>• Scheduled quarter-hourly arrival and departure demand<br>• Quarter-hourly airport capacity: effective Airport Acceptance Rate (AAR), Airport Departure Rate (ADR)<br>• Observed airport performance: quarter-hourly effective arrival and departure counts, on time arrival and departure percentage<br>• Times: Hour, day of week, month<br>• Delays: Quarter-hour average arrival delays (based on flight plan) and departure delays[4]<br>• On-time Percentage: Quarter-hour arrival and departure on-time percentage statistics |
| **Traffic Management Initiatives (TMIs)** | Layered TMIs | Commonly used TMI identified in the selected regions: GDPs, GSs, AFPs, Reroutes |

---

[4] In ASPM, DLAFPARRA is the quarter-hour average arrival delays (based on flight plan), and DLAFPOFFA is the quarter-hour average arrival wheels-off delays.

## A. Weather Data

The weather data we used is from CIWS, which provides gridded storm severity information in a 1km x 1km resolution [4]. Fig. 2 shows an example of CIWS data samples we processed. For each grid, we fetched two data items: Vertically Integrated Liquid (VIL) and Echo Top (ET). VIL is related to reflectivity, the VIL values are mapped to the corresponding six levels (1 to 6) to present weather severity. And ET's unit is feet and is mapped to fifteen levels (0-14) according to our defined criteria.

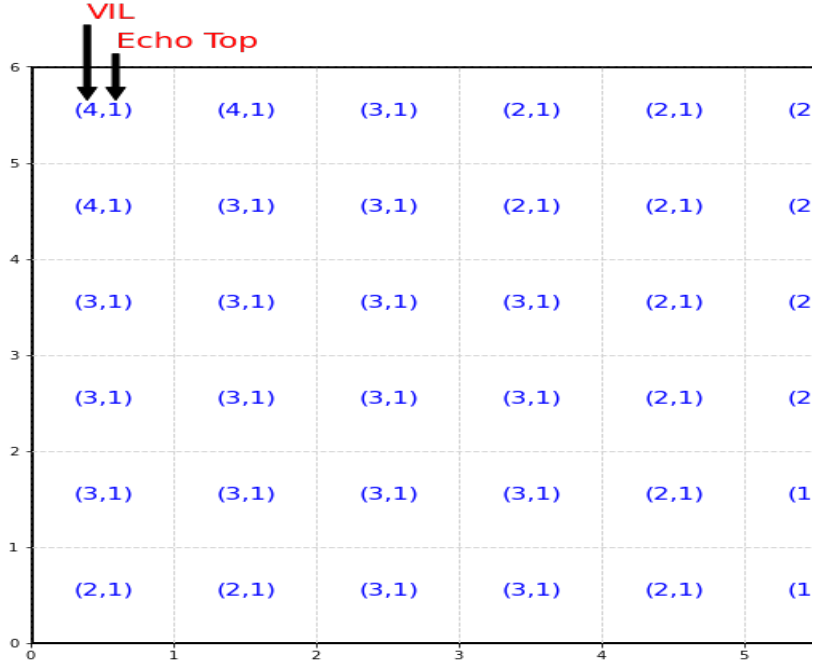

**Fig. 2 CIWS data examples.**

## B. Traffic and Performance Data

The FAA's ASPM dataset is a well-known data source for airport performance review and reporting [16]. It integrates multiple data sources together and provides information about flights arriving or departing an airport. The ASPM dataset provides aggregate quarter-hourly demand, delays, airport capacity, on-time statistics, etc., so it serves the main source of traffic and performance data for the delay prediction model. Table 1 lists the features retrieved from the APSM dataset.

## C. Traffic Management Initiatives (TMIs) Data

The third dataset for delay prediction is the usage of TMI (e.g., Ground Delay Program (GDP), Ground Stop (GS), Airspace Flow Program (AFP), and reroutes) from Traffic Flow Management System (TFMS) National Traffic Management Log (NTML) messages, and it is then processed to our required format by our scripts. Due to the textual nature of TMI data, they are treated as individual categorical variables and transformed into a binary format and mapped into quarter-hour bins, i.e., 1 if a specific TMI used in that quarter hour, and 0 otherwise. Below is the list of specific TMIs used in the prediction model:

- TMIs for enroute airspace: AFP of FCAA08, FCABW1, FCADC1, FCADC7, FCAID1, FCAN92, FCAOB1, FCAOB3, and FCAOB6, respectively
- Any reroute regarding ZNY inbound traffic
- Any reroute regarding ZNY outbound traffic
- TMIs for Individual Airports
- GDP
- GS
- Any reroute regarding airport inbound traffic
- Any reroute regarding airport outbound traffic

# V. Modeling Methods

After consulting subject matter experts, the study area of NEC is bounded in a rectangular box defined as the orange box in Fig. 3. The CIWS data dimension of weather data for NEC area is 960 by 1072 by 2, which is challenging to ingest as-is by any prediction model.

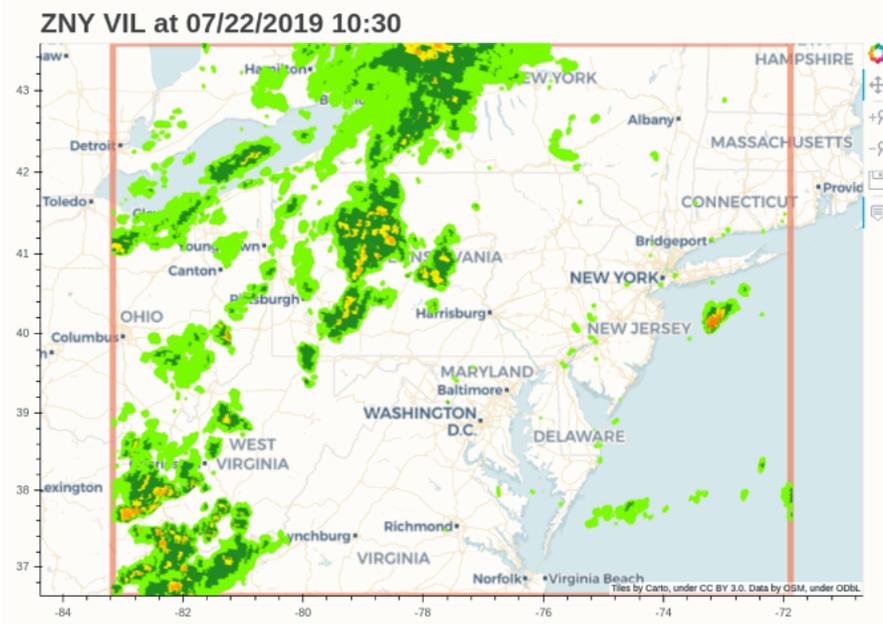

**Fig. 3 Boundary defined North-east corridor (NEC) study area.**

With consideration of high dimensional CIWS weather data, we designed a two-step learning architecture to support delay prediction (see Fig. 4). The first step is to learn a low-dimensional data representation for CIWS data with autoencoder. And then the second step is to predict delay with a Transformer-based technique. Next, more details about these two steps are discussed.

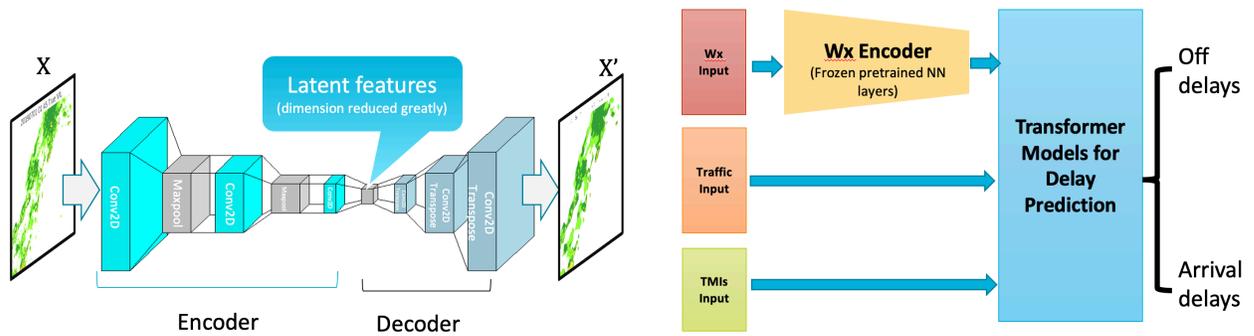

**Fig. 4 Delay prediction learning architecture**

### Step 1: Encoding Weather Data Encoding

To effectively incorporate the high-dimensional CIWS data, we employed a generative SSL technique – autoencoder [17], to learn an effective representation for CIWS data. Such a data representation is expected to be in a much lower dimension but to still contain enough characteristics to differentiate patterns or characteristics to support downstream applications. In addition, the learned representation can also make the downstream applications easier.

With consideration that CIWS data format are like image data format, we built our autoencoder with a 2D convolution neural network (CNN) (e.g., Conv2D in TensorFlow [18]). We used Keras and TensorFlow (version 2.4.0) to develop our autoencoder models. For autoencoders, the shape of inputs and output should be same. Therefore, we selected Conv2D to build our encoder, and Conv2DTranspose to build our decoder. Fig. 5 is the autoencoder model architecture built for this effort and depicted by TensorFlow. It details the size of each neural network layers and the depth of the whole CNN. Within the CNN encoder, a series of Conv2D layers reduces the shape of the input image from 960x1072x2 to 29x32x16. Then, through the CNN decoder, a series of Conv2DTranspose layers expands the dimension back to the original shape 960x1072x2. The autoencoder model is to be trained with the goal to minimize the numerical differences between the input of the encoder and the output of the decoder. During the training process, the encoder will be trained to reduce the dimension of the original image but keep enough information so that the decoder can produce an image that matches the original. Therefore, once the training is done, the encoder can be used to obtain meaningful data representation of a CIWS image in a much lower dimension, which in this case is 29x32x16. To further reduce the dimension, we applied a common practice of deep learning, global average method, to get 16-dimention feature vector, which can then be used to support downstream AI applications.

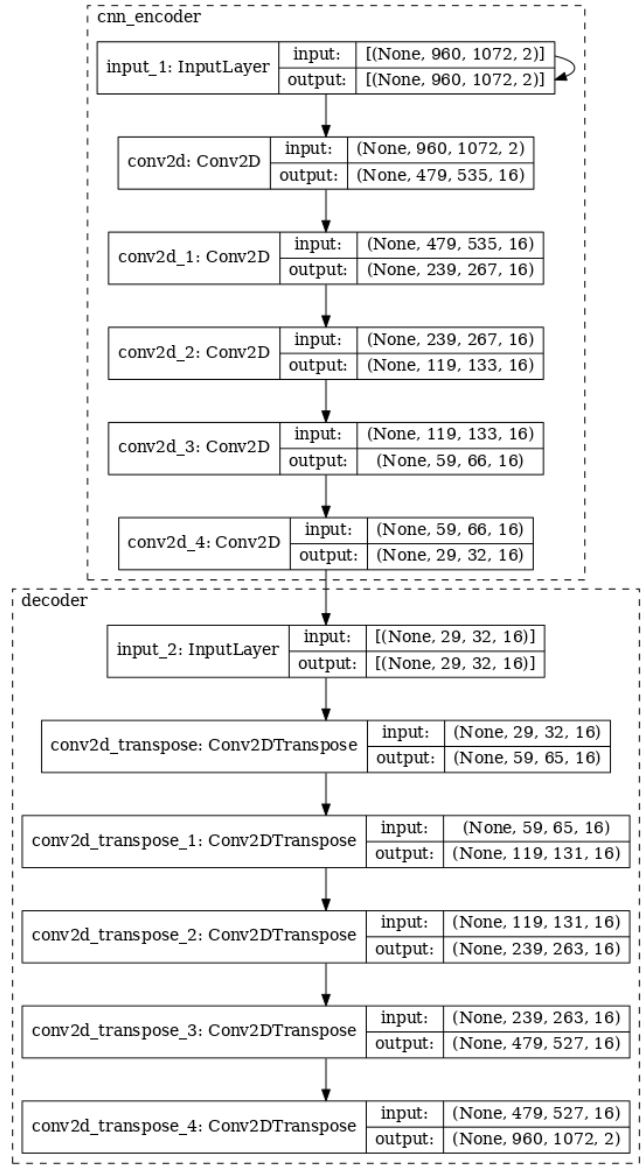

**Fig. 5 CIWS autoencoder implementation network structure.**

**Step 2: Delay Prediction Model with Temporal Fusion Transformer (TFT)**

Flight delay prediction is a multi-step, multi-variable times series prediction problem, which is challenging due to the heterogeneous input variables (i.e., weather data, traffic data, and TMIs data), the unknown interactions among the inputs, temporal dynamics etc. Recently, "Transformer" introduced by Google in 2017 [7] has become de-facto standard in many sequence-to-sequence applications (e.g., language translation, time series prediction). Inspired by transformer's versatility and superior performance, we adopted Temporal Fusion Transformer (TFT) [13] for our delay prediction. TFT has been proven to beat Amazon's DeepAR by 36–69% in benchmarks [12] and achieve significant performance improvement over a wide range of real-world time series forecasting problems [13].

TFT is an innovative attention-based neural network architecture, and its supremacy comes from its internal design of multiple components. Fig. 6 shows TFT architecture. The major components of TFT are explained as follows:

- **Variable Selection:** It learns the importance of the input variables for the response variables at each time step.
- **Gating mechanisms**: It can filter out any unused components of the architecture, providing adaptive depth and network complexity to accommodate various types of datasets and scenarios.
- **Static covariate**: It integrates static features into the network, through encoding of context vectors to condition temporal dynamics.
- **Temporal processing**: It learns both long- and short-term temporal relationships from both observed and known time-varying inputs. A sequence-to-sequence layer is employed for local processing, while long-term dependencies are captured using a novel interpretable multi-head attention block.

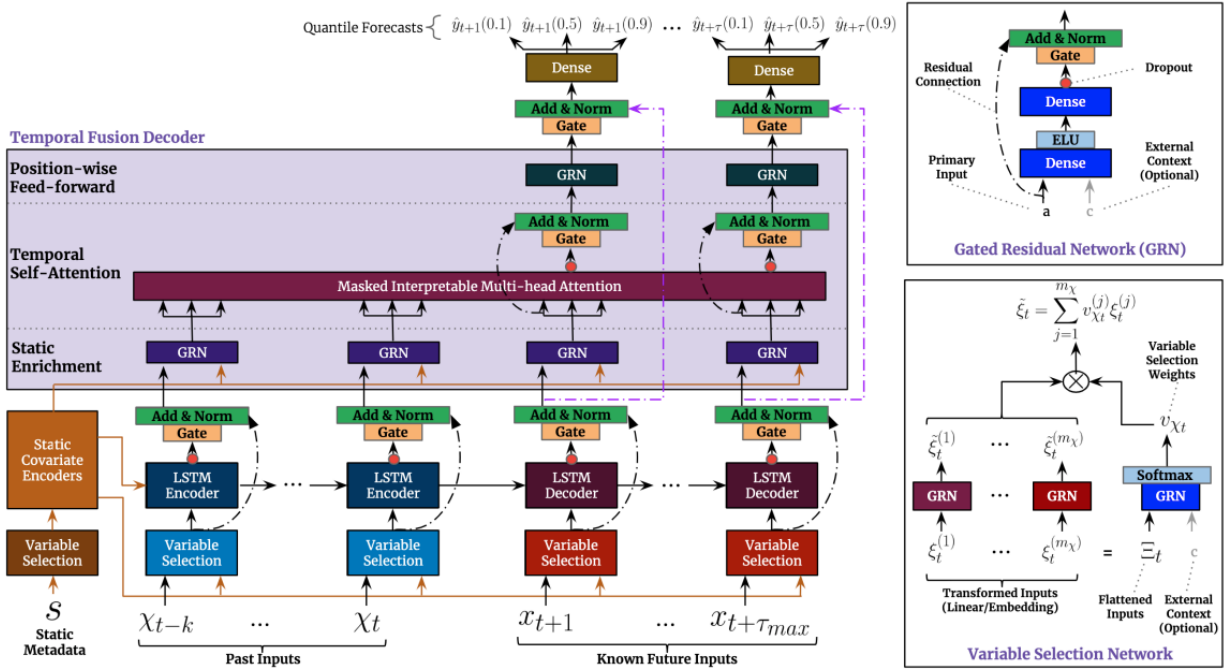

**Fig. 6 TFT architecture [13].**

The inputs and the outputs of our TFT model need to be prepared from the sources specified in Section III. As shown in Fig. 6. Table 2 lists the detailed inputs to TFT for delay prediction

**Table 2 Detailed inputs for each category to our TFT delay prediction model**

| Variable Types | Variables |
| --- | --- |
| **Static Metadata** | Airport ID (i.e., LGA, JFK, EWR, PHL) |
| **Observed Inputs** | Actual arrivals count, Actual departures count, past average arrival delay, past average departure delay, past arrival on-time percentage, past departure on-time percentage |
| **Known Inputs** | Arrival demand, Departure demand AAR, ADR, AAREFF, implementation status of 15 TMIs, hour, time of day, month |

The output (or prediction target) for TFT is future average departure (wheels-off) delays and future average arrival delays of each airport in 15 minutes bin. Because the average delays per quarter hour could be noisy when few flights were present during off-peak periods, therefore, moving average was taken to smooth out short-term fluctuations and highlight longer-term trends or cycles. Thus, the moving average of delays were the target for prediction. Fig. 7 shows an example of flight plan departure wheel-off delays (blue line-DLAFPOFFA per ASPM definition) compared to its moving average (orange line).

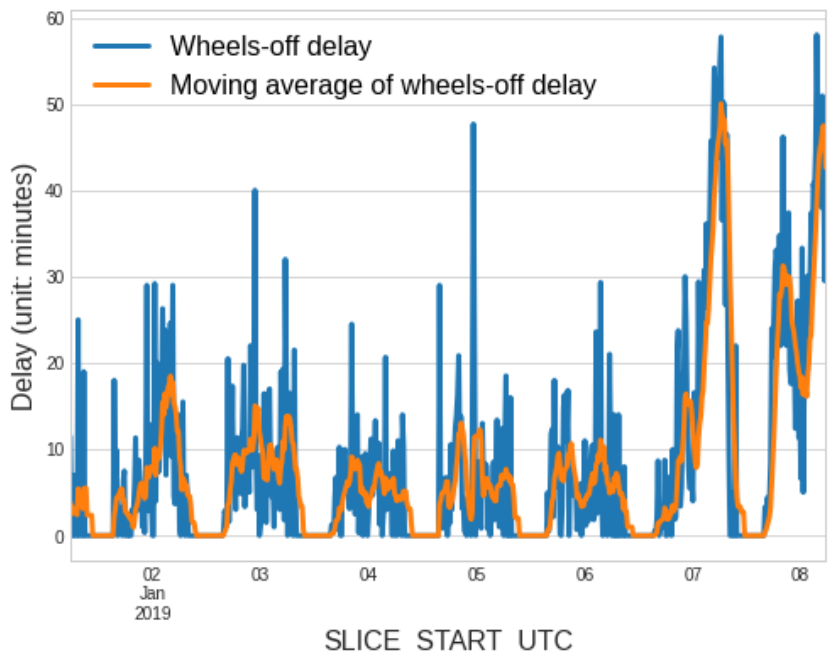

**Fig. 7 Example of moving average of departure wheels-off delays.**

## VI. Results

In this section, we will present our encoding results and delay prediction results for NEC four airports (LGA, JFK, EWR, and PHL).

### A. CIWS Data Encoding Results

We trained our autoencoder for CIWS weather data with 5 months of data (05/2019-09/2019) and tested with 10/2019 data. The encoder can compress the original image to 0.7% of original size and still can recover back in good fidelity (see Fig. 8) with absolute errors less than 3 categories for both of VIL and ET, respectively. When training is done, the encoder model is deployed to compress CIWS weather data to support various types of downstream AI applications. Currently, we have employed the encoder to get weather feature vectors for two years of CIWS data (2019-2020).

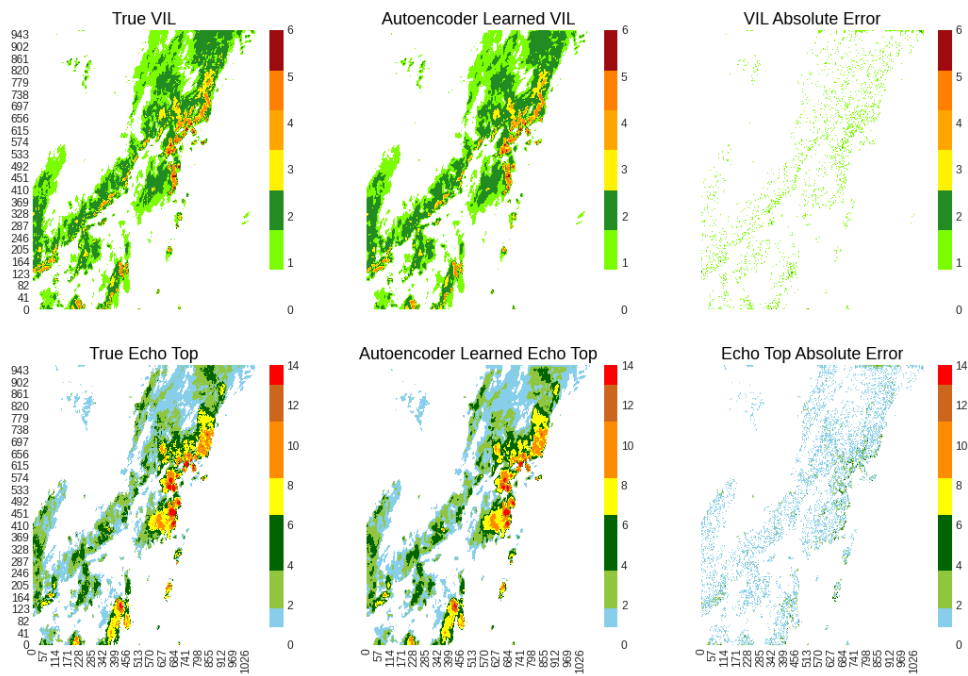

**Fig. 8 An example of CIWS autoencoding results.**

## B. Delays Prediction Results

For TFT, we trained the model with whole year of 2019, and then tested its performance with 01/2020 data. The look-ahead time $\tau_{max}$ for TFT delay prediction was set to four hours. And the time lag *k* was learned from the training data, which is one hour in our case.

Fig. 9 shows the model performance for four airports on the testing dataset of 01/2020, measured by mean squared error (mse). It demonstrates that arrival delay prediction error is smaller than that of departure delay. And the model performs the best at LGA airport.

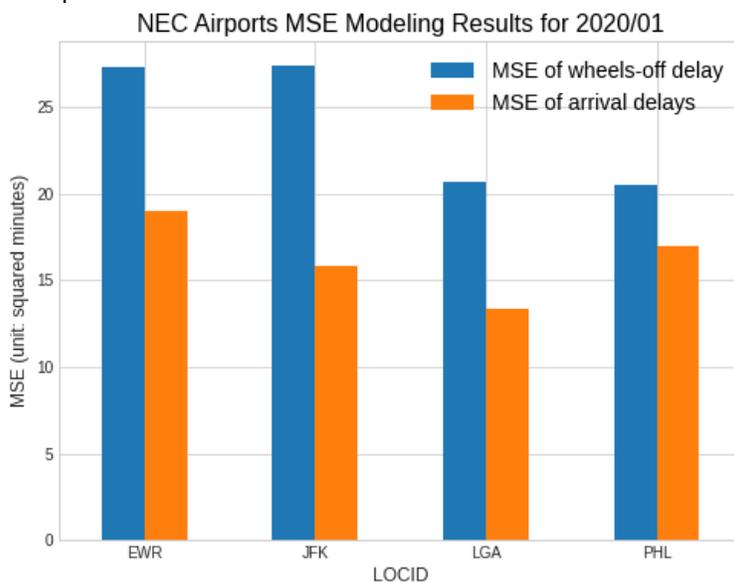

**Fig. 9 Four-airport delay prediction performance comparison on the testing dataset.**

To understand the temporal differences of the prediction performance, Fig. 10 and Fig. 11 show the prediction results of both departure and arrival delays, respectively at LGA for selected testing days in January 2020, which was never seen in the training process. The comparison of actual (solid line) with predicted (dashed line) values clearly demonstrates that the TFT model can capture the upward and downward trends of delays. The small values of mse and mean absolute error (mae) show the low error margins achieved by the TFT model. Fig. 12 to Fig. 17 show the testing results at the other three airports, which were predicted all at once with the LGA delays.

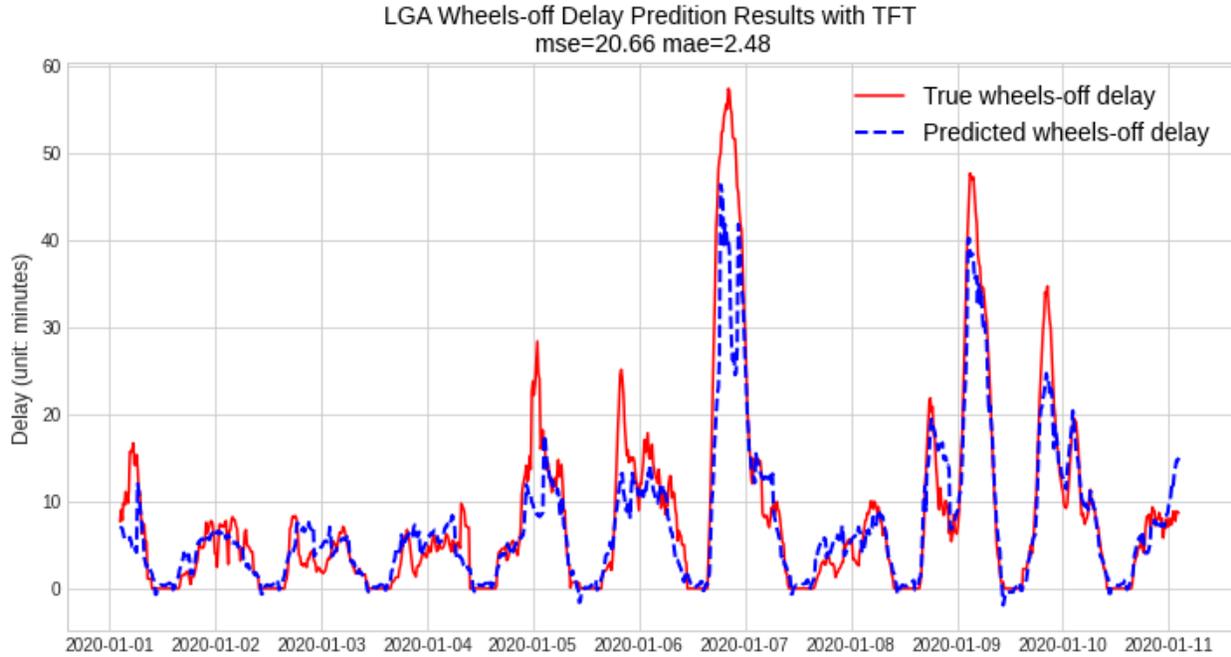

**Fig. 10 An example of departure delays prediction for LGA.**

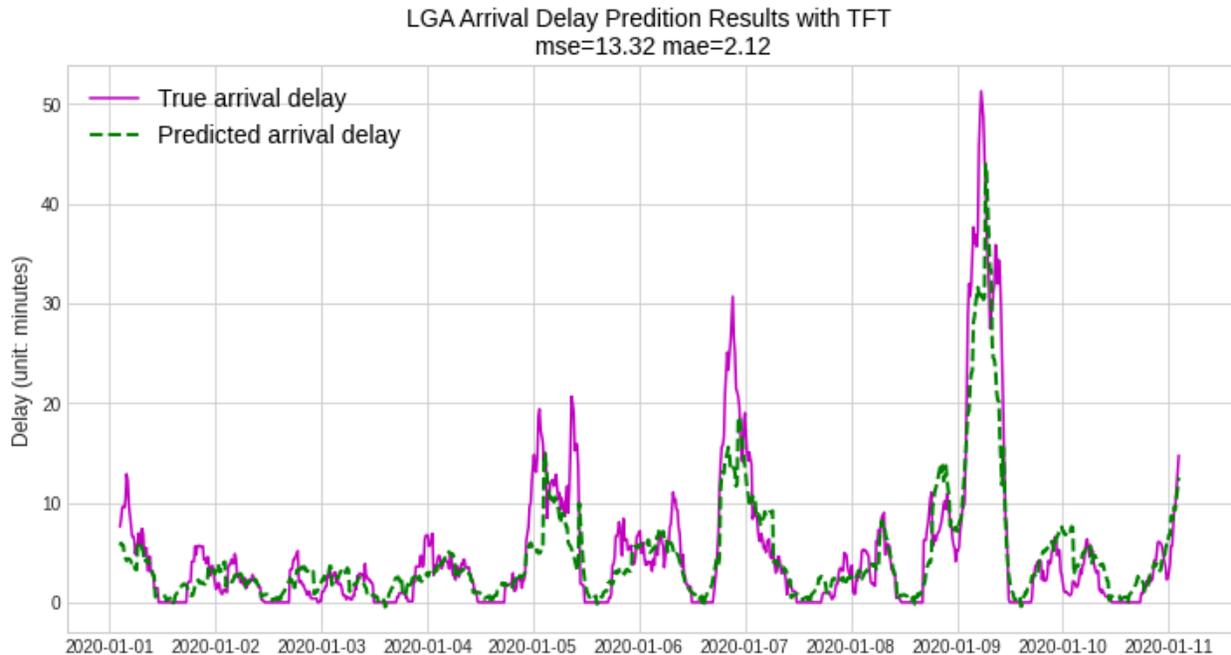

**Fig. 11 An example of arrival delays prediction for LGA.**

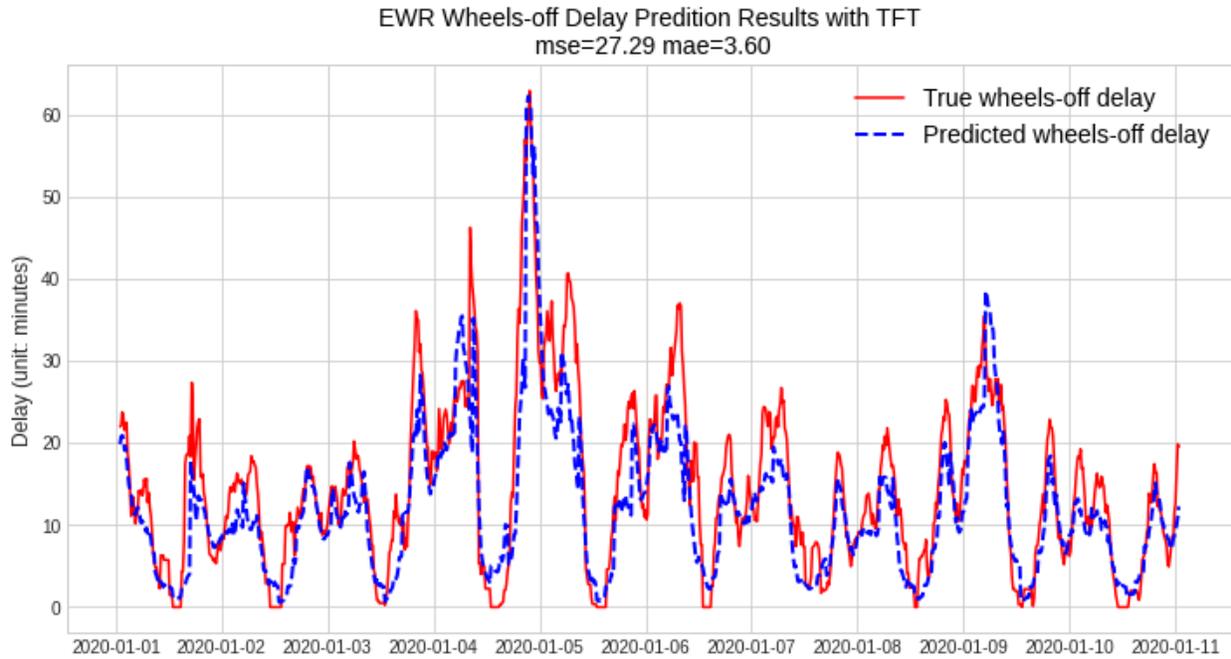

**Fig. 12 An example of departure delay prediction for EWR.**

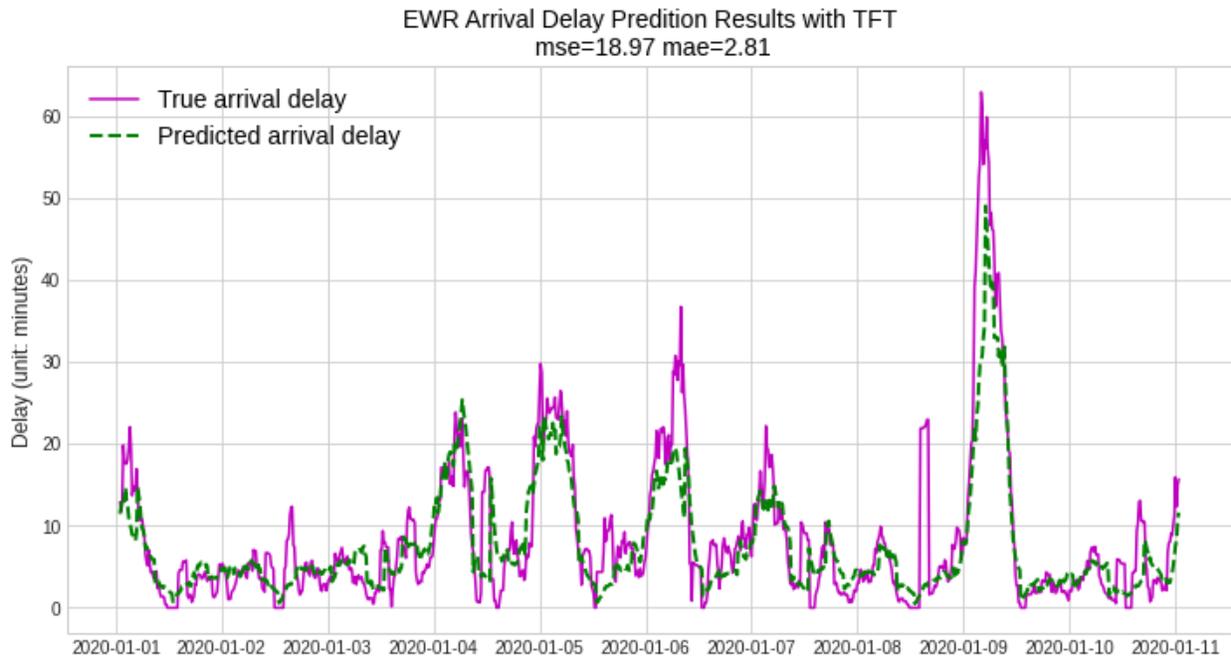

**Fig. 13 An example of arrival delay prediction for EWR.**

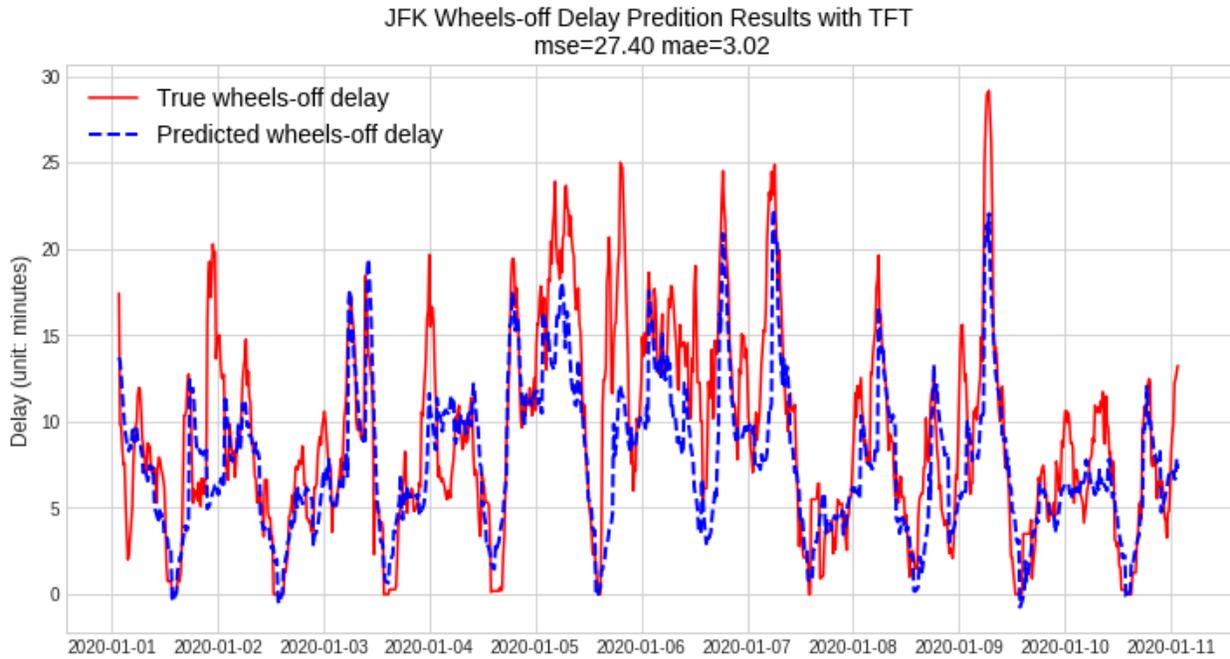

**Fig. 14 An example of departure delay prediction for JFK.**

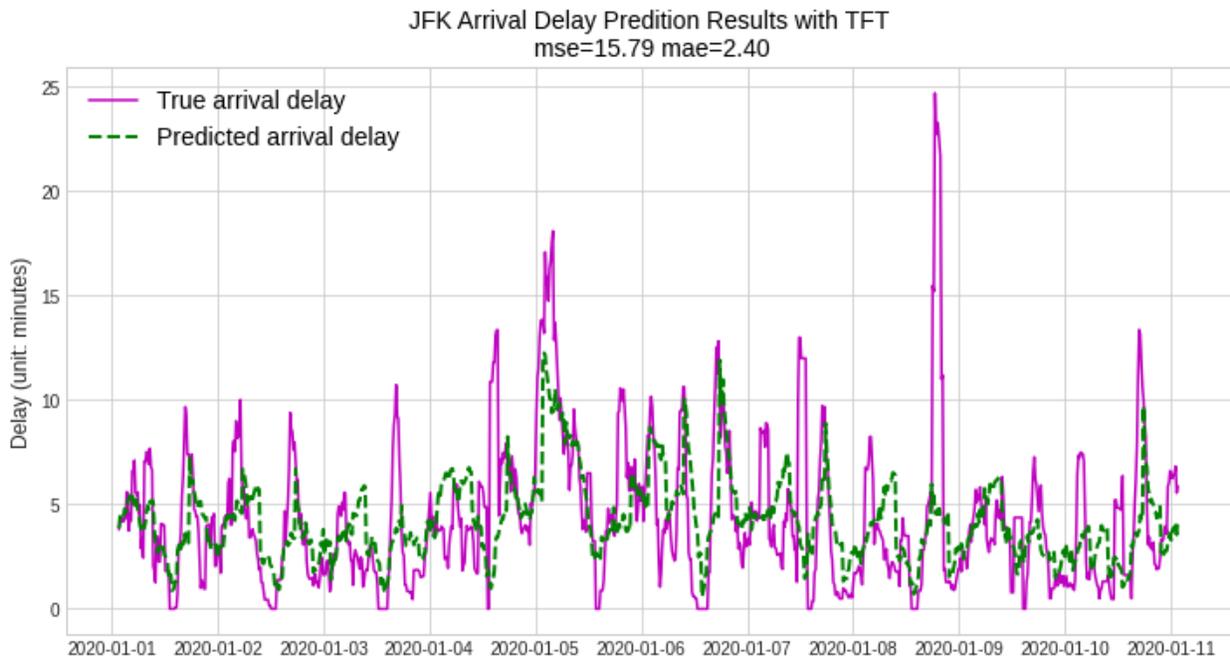

**Fig. 15 An example of arrival delay prediction for JFK.**

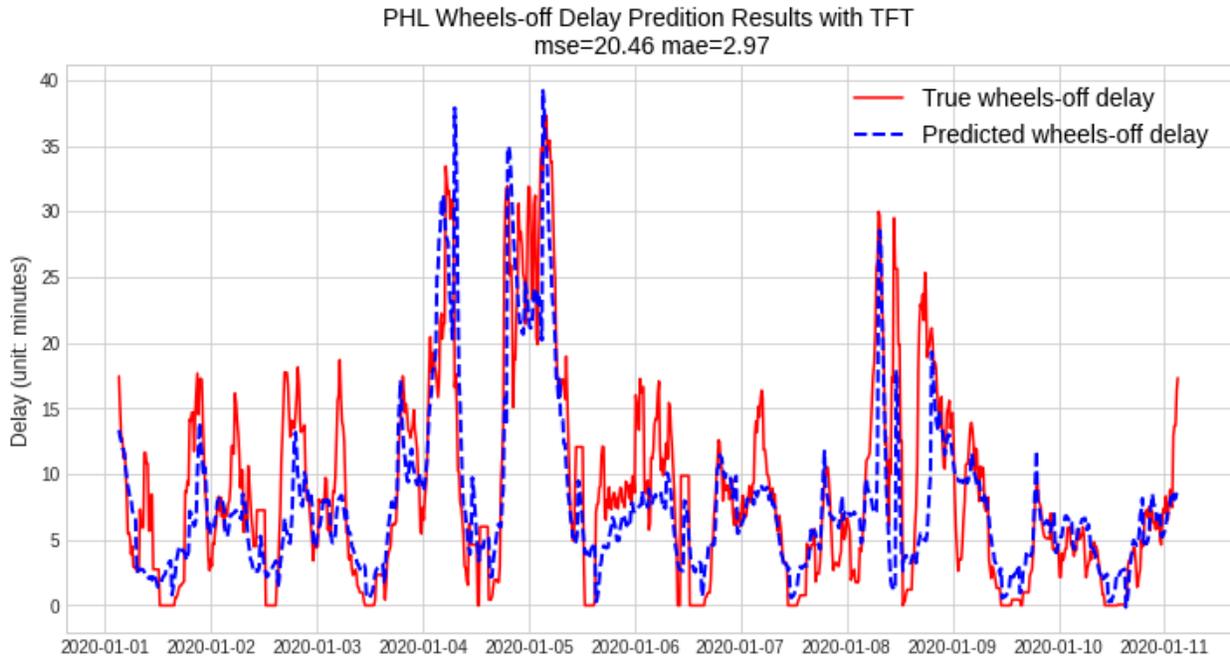

**Fig. 16 An example of departure delay prediction for PHL.**

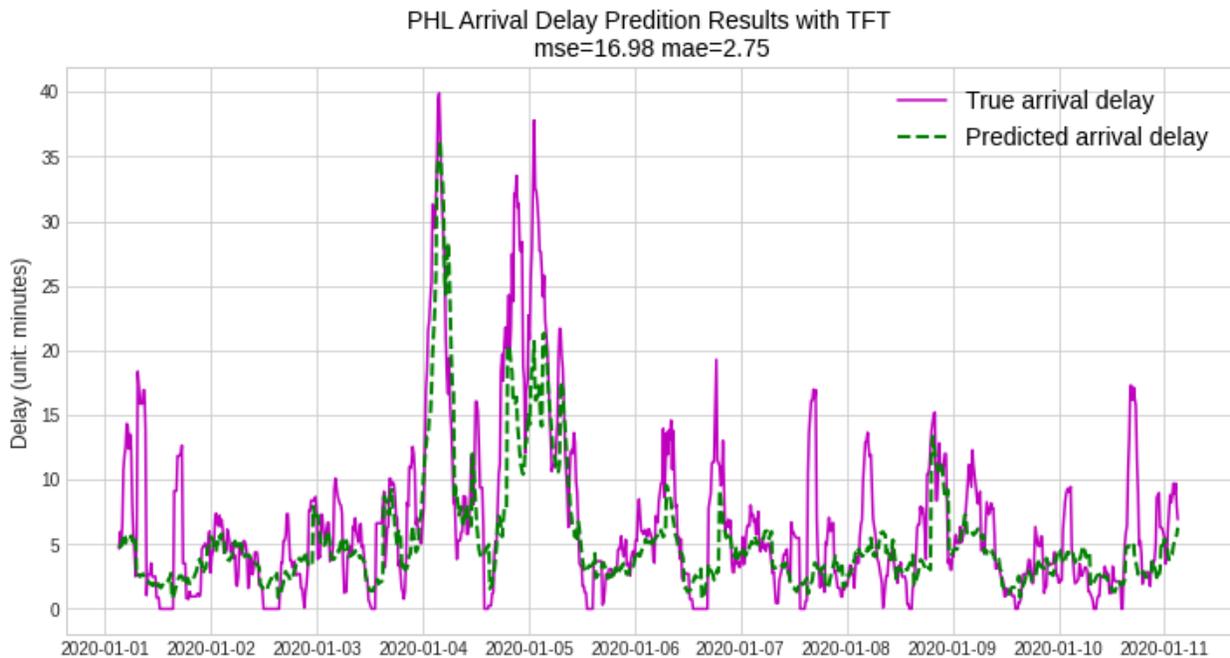

**Fig. 17 An example of arrival delay prediction for PHL.**

Note that, in Fig. 10 to Fig. 17, the testing data had large spikes of actual delays that were all due to severe winter weather, and the prediction errors tend to be larger in some time periods on those days at JFK and PHL. Possible causes are also examined. First, we have more good winter weather days than bad winter weather days in the training dataset. Due to the imbalanced nature of the data, the model would perform better (in terms of error minimization) on the non-severe weather days. Second, the severe weather days were not exactly similar to each other, and the TMIs applied to those days were also different. Those combining factors have made the model difficult to predict and caused large prediction errors for certain periods during bad weather days. Despite all those challenges, the mean absolute errors (MAE) are below 4 minutes, and the upward/downward trends of the predicted delays are closely following the actual ones. Thus, the utility of using the model to enable traffic managers to mitigate potential weather impact should be further researched.

### C. TFT Model Interpretability

The first nice feature of TFT is its interpretability of time dynamics. For example, Fig. 18 shows the attention score relative to time index. A higher attention score means a higher contribution from that time step to the prediction outputs. It should be noted that time index - 1 represents t-1 timestamp's importance for the predictions. As the figure clearly demonstrates, the nearest time steps play bigger roles, which is in expectation.

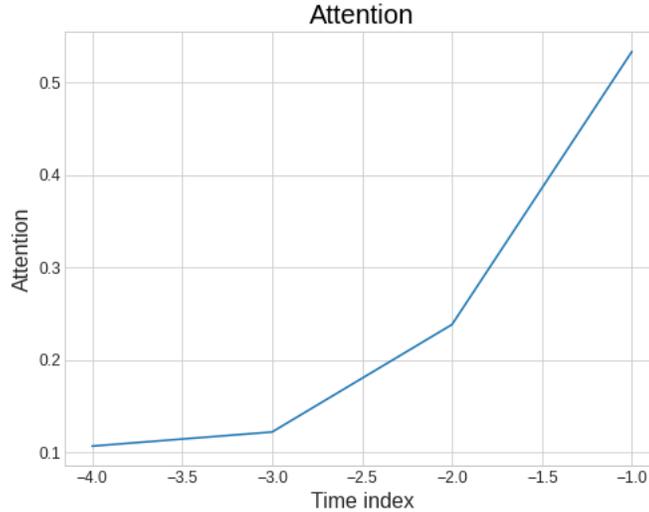

**Fig. 18 Attention score by time index.**

The second interpretability feature of TFT is the importance of each input variable for the response variables. For example, from, the relative importance of the input variables for delay prediction can be identified. For the TFT encoder inputs, one Flow Constrained Area (FCA) and one weather element are top 2 factors. For the TFT decoder input variables, FCAs and scheduled demands are influential factors. For the static variable, airport ID plays a key role in prediction.

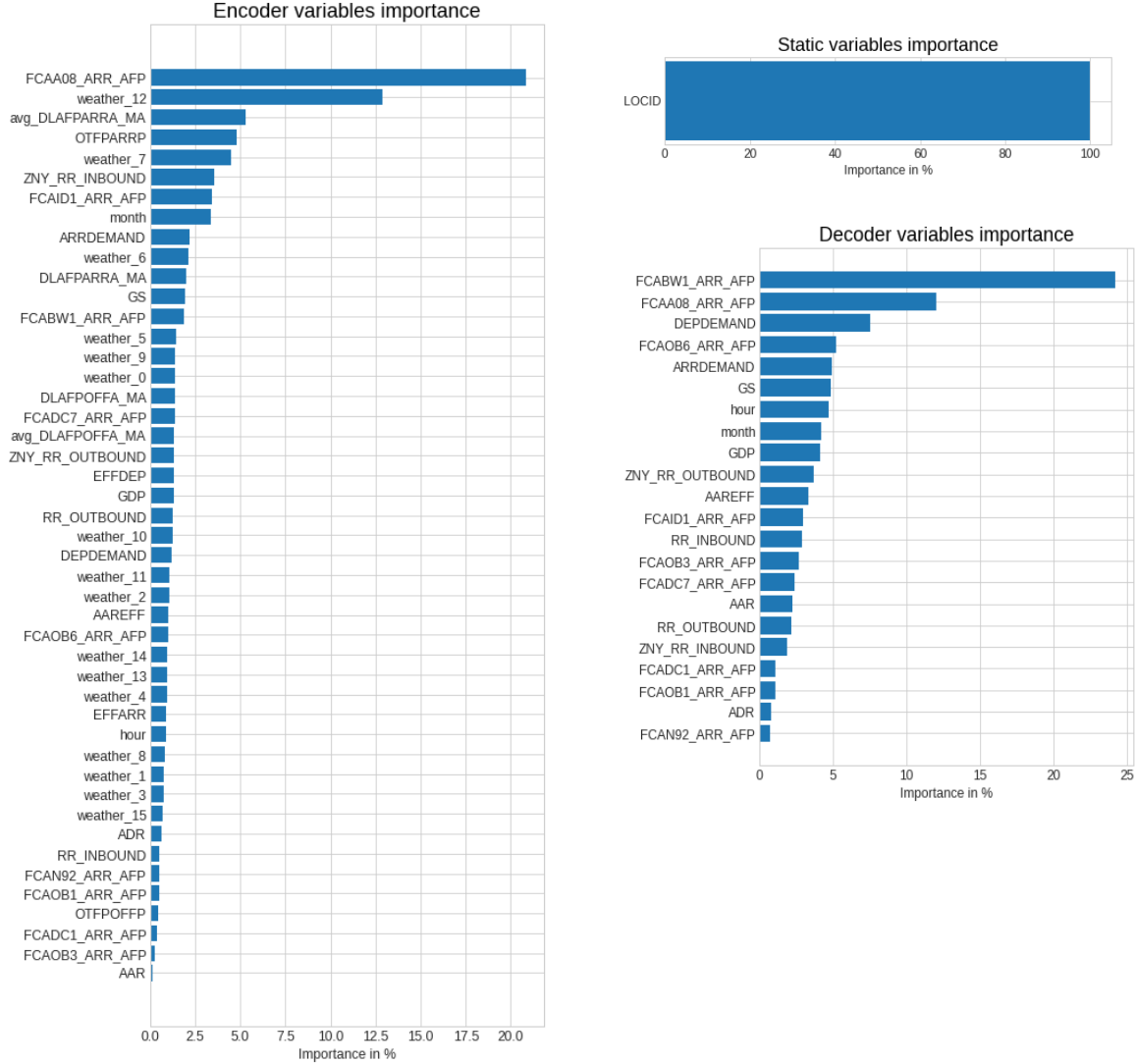

(a) Encoder variables importance  (b) Static and decoder variables importance

**Fig. 19 Importance of variables in delay prediction.**

Unlike the computer vision or NLP field, aviation domain does not have a recognized machine learning benchmarking dataset for comparing flight delay prediction models. Therefore, it is difficult to compare different models published in various papers due to their different study subjects, different datasets, and forecasting horizon. A few references suggested that the equivalent mse of departure delay prediction was between 156 [19] and 529 [20]. As a comparison, the mse of TFT departure delay prediction is between 20 and 27, which is significantly smaller than those of prior research. Nonetheless, determining if these errors are sufficiently small enough may depend on the operational use cases and feedback from field evaluations.

## VII.  Summary

We applied modern AI techniques to develop a regional multi-airport delay prediction model. An autoencoder was first applied to learn an effective weather data representation and support downstream AI applications. Then, a transformer-based model, TFT, was adopted to build a multi-airport delay prediction model, which can predict departure and arrival delays simultaneously for four major airports (i.e., PHL, LGA, JFK, and EWR). The TFT model can handle heterogeneous inputs, such as weather, traffic, and TMIs data. The performance of the trained model on the testing dataset did not degrade, which indicated model generality.

For next steps, to further validate the model performance, a field test with live data ingestion is needed. The feedback will inform model refinement, prototype design, and operationalization requirements. Secondly, the model can be further extended to support TMI planning by predicting the performance with or without a TMI. With such a sensitivity analysis, the model can support the identification of which TMIs are effective in delay mitigation. Thirdly, the TFM model performance could be further improved with additional input features, such as airport wind, ceiling, and runway configurations etc. Lastly, the probabilistic nature of the TFT outputs should be explored, as the current outputs are based on the median of the distribution of the predicted delays. Leveraging the model's quantile output capabilities (e.g., 25th percentiles, median, 75th percentile) across multi-time horizons can help the user understand the prediction uncertainty and build the trust on this AI application.

## Acknowledgments

We thank the following MITRE colleagues: Mike Robinson, Bill Bateman, Lixia Song, Mike Klinker, Erik Vargo, Steve Zobell, and Panta Lucic for the support, valuable discussions, and insights.

## NOTICE

## Bibliography

[1] Federal Aviation Administration, "Fact Sheet – Inclement Weather," 24 March 2021. [Online]. Available: https://www.faa.gov/news/fact_sheets/news_story.cfm?newsId=23074. [Accessed 1 July 2021].

[2] FAA, "Cost of Delay Estimates," 2019. [Online]. Available: https://www.faa.gov/data_research/aviation_data_statistics/media/cost_delay_estimates.pdf. [Accessed 14 Jan 2020].

[3] FAA, "Weather Delay," [Online]. Available: https://www.faa.gov/nextgen/programs/weather/faq/. [Accessed 14 Jan 2020].

[4] D. Klingle-Wilsonk and J. Evans, "Description of the Corridor Integrated Weather System (CIWS) Weather Products," 1 Aug 2005. [Online]. Available: https://www.ll.mit.edu/sites/default/files/publication/doc/2018-12/Klingle-Wilson_2005_ATC-317_WW-15318.pdf. [Accessed 14 Jan 2020].

[5] NOAA, "The Rapid Update Cycle (RUC)," NOAA, [Online]. Available: https://ruc.noaa.gov/ruc/.

[6] Meteorological Satellite Center of JMA, "Meteorological Satellite Center of JMA," Meteorological Satellite Center of JMA, [Online]. Available: https://www.data.jma.go.jp/mscweb/data/himawari/.

[7] A. Vaswani, N. Shazeer, N. Parmar, J. Uszkoreit, L. Jones, A. N. Gomez, L. Kaiser and I. Polosukhin, "Attention is all you need," 6 Dec 2017. [Online]. Available: https://arxiv.org/abs/1706.03762. [Accessed 15 Jan 2020].

[8] J. J. Rebollo and Hamsa Balakrishnan, "Characterization and prediction of air traffic delays," *Transportation Research Part C,* vol. 44, pp. 231-241, 2014.

[9] B. Yua, Z. Guoa, S. Asianb, H. Wang and G. Chen, "Flight delay prediction for commercial air transport: A deep learning approach," *Transportation Research Part E,* no. 125, pp. 203-221, 2019.

[10] M. Zoutendijk and M. Mitici, "Probabilistic Flight Delay Predictions Using Machine Learning and Applications to the Flight-to-Gate Assignment Problem.," *Aerospace,* vol. 8, no. 6, 2021.

[11] A. Kalliguddi and A. Leboulluec, "Predictive Modeling of Aircraft Flight Delay," *Universal Journal of Management,* vol. 5, no. 10, pp. 485-491, 2017.

[12] J. Bertiner, "Introducing PyTorch Forecasting," 19 Sep 2020. [Online]. Available: https://towardsdatascience.com/introducing-pytorch-forecasting-64de99b9ef46. [Accessed 12 Apr 2020].

[13] B. Lim, S. O. Arik, N. Loeff and T. Pfister, "Temporal Fusion Transformers for Interpretable Multi-horizon Time Series Forecasting," 27 Sep 2020. [Online]. Available: https://arxiv.org/abs/1912.09363. [Accessed 16 Jan 2021].

[14] C. Fan, Y. Zhang, Y. Pan, X. Li, C. Zhang, R. Yuan, D. Wu, W. Wang, J. Pei and H. Huang, "Multi-Horizon Time Series Forecasting with Temporal Attention Learning," in *KDD*, Anchorage, AK, 2019.

[15] N. Wu, B. Green, X. Ben and S. O'Banion, "Deep Transformer Models for Time Series Forecasting: The Influenza Prevalence Case," 13 Jan 2020. [Online]. Available: https://arxiv.org/pdf/2001.08317.pdf. [Accessed 1 June 2020].

[16] "Aviation System Performance Metrics (ASPM)," 26 5 2020. [Online]. Available: https://aspmhelp.faa.gov/index.php/Aviation_Performance_Metrics_%28APM%29.

[17] G. Hinton and R. Salakhutdinov, "Reducing the Dimensionality of Data with Neural Networks," *Science,* vol. 313, no. 5786, pp. 504-507, 2006.

[18] Google, "Tensorflow API Documentation," [Online]. Available: https://www.tensorflow.org/api_docs. [Accessed 1 Jun 2020].

[19] A. Kalliguddi and A. Leboulluec, "Predictive Modeling of Aircraft Flight Delay," *Universal Journal of Management,* vol. 5, no. 10, pp. 485-491, 2017.

[20] J. Rebollo and H. Balakrishnan, "Characterization and Prediction of Air Traffic Delays," *Transportation Research Part C,* vol. 44, pp. 231-241, 2014.

[21] Z. Wang, S. Lance and J. Shortle, "Improving the Nowcast of Unstable Approaches," in *Integrated Communications, Navigation, Surveillance (ICNS)*, Dullas,VA, 2016.

[22] A. Maheshwari, N. Davendralingam and D. DeLaurentis, "A Comparative Study of Machine Learning Techniques forAviation Applications," in *Aviation Technology, Integration, and Operations Conference*, Altalanta, GA, 2018.

[23] Y. J. Kim, S. Choi, S. Briceno and D. Mavris, "A deep learning approach to flight delay prediction," in *IEEE/AIAA 35th Digital Avionics Systems Conference (DASC)*, Sacramento, CA, 2016.

[24] H. Lee and W. Malik, "Taxi-Out Time Prediction for Departures at Charlotte Airport Using Machine Learning Techniques," in *16th AIAA Aviation Technology, Integration, and Operations Conference*, Washington, D.C., 2016.

[25] L. Weng, "From Autoencoder to Beta-VAE," 12 Aug 2018. [Online]. Available: https://lilianweng.github.io/lil-log/2018/08/12/from-autoencoder-to-beta-vae.html. [Accessed 10 Oct 2020].